\documentclass[pmlr]{jmlr}% new name PMLR (Proceedings of Machine Learning)
\usepackage{longtable}% for long tables

\usepackage{adjustbox}

\usepackage{booktabs}
\usepackage[load-configurations=version-1]{siunitx} % newer version
 \usepackage[normalem]{ulem}
\usepackage{graphicx}
\usepackage{subcaption}
\usepackage{makecell}

\theorembodyfont{\upshape}
\theoremheaderfont{\scshape}
\theorempostheader{:}
\theoremsep{\newline}

\jmlrvolume{85}
\jmlryear{2018}
\jmlrworkshop{Machine Learning for Healthcare}

\title{A Domain Guided CNN Architecture for Predicting Age from Structural Brain Images}

\author{\Name{Pascal Sturmfels} \Email{psturm@umich.edu} 
       \addr Department of Electrical Engineering\\  University of Michigan \hspace*{\fill} and Computer Science\\
       Ann Arbor, MI
       \AND
       \Name{Saige Rutherford} \Email{saruther@umich.edu} 
       \addr Department of Psychiatry\\
       University of Michigan\\
       Ann Arbor, MI
       \AND
       \Name{Mike Angstadt} 
       \Email{mangstadt@med.umich.edu} 
       \addr Department of Psychiatry\\
       University of Michigan\\
       Ann Arbor, MI
       \AND
       \Name{Mark Peterson} \Email{mdpeterz@med.umich.edu} 
       \addr Department of Physical Medicine\\
       University of Michigan \hspace*{\fill} and Rehabilitation\\
       Ann Arbor, MI
       \AND
       \Name{Chandra Sripada}
       \Email{sripada@umich.edu} 
       \addr Department of Psychiatry\\
       University of Michigan\\
       Ann Arbor, MI
       \AND
       \Name{Jenna Wiens} \Email{wiensj@umich.edu} 
       \addr Department of Electrical Engineering\\  University of Michigan \hspace*{\fill} and Computer Science\\
       Ann Arbor, MI
       }

\firstpageno{1}

\begin{document}

% \title{Adapting CNNs for Neuroimaging Tasks by Exploiting Brain Structure: A Case Study on Predicting Age}`

%NOTES
%MAKE CLEAR HOW WE THINK OF FILTERS,  e.g. as 4D tensors over the spatial dimensions and the input channel dimension. 

\maketitle

\begin{abstract}
Given the wide success of convolutional neural networks (CNNs) applied to natural images, researchers have begun to apply them to neuroimaging data. To date, however, exploration of novel CNN architectures tailored to neuroimaging data has been limited. Several recent works fail to leverage the 3D structure of the brain, instead treating the brain as a set of independent 2D slices. Approaches that do utilize 3D convolutions rely on architectures developed for object recognition tasks in natural 2D images. Such architectures make assumptions about the input that may not hold for neuroimages. For example, existing architectures assume that patterns in the brain exhibit translation invariance. However, the meaning of a pattern may differ depending on its location in the brain. There is a need to explore novel architectures that are tailored to neuroimaging data. We present two simple modifications to existing CNN architectures to better learn from structural neuroimaging data. Applied to the task of brain age prediction, our network achieves a mean absolute error (MAE) of 1.4 years and trains 30\% faster than a CNN baseline that achieves a MAE of 1.6 years. Our results suggest that lessons learned from developing models on natural images may not directly transfer to neuroimaging tasks. Instead, there remains a large space of unexplored questions regarding model development in this area, whose answers may differ from conventional wisdom.

\end{abstract}

\section{Introduction}\label{sec:Introduction}
Convolutional neural networks (CNNs) have achieved state-of-the-art performance across a variety of tasks involving natural images, including object and action recognition. Convolutional layers come with a host of benefits, including a more efficient usage of parameters compared to standard neural networks and the ability to learn patterns at multiple scales. In addition, convolutional layers allow the network to learn spatially invariant relationships between input and output. This is particularly useful in vision tasks involving natural images, in which an object's identity is often independent from its location in the image.

Researchers from the neuroimaging community have recently begun exploring the utility of CNNs applied to publicly available brain-image datasets, predicting neurological conditions like Alzheimer's \citep{sarraf2016deepad} and autism \citep{anirudh2017bootstrapping}, as well as age \citep{cole2017predicting} and survival time of high-grade glioma patients \citep{nie20163d}. Although structural brain scans and natural images share similarities, there are several key differences that make adapting CNNs to brain images non-trivial. 

Much of the existing work on CNNs has focused on 2-dimensional (2D) images, but structural MRI scans of the human brain are 3-dimensional (3D). Some previous work has treated 3D volumes as a set of independent 2D slices \citep{sarraf2016deepad,farooq2017deep}, but doing so fails to fully leverage the spatial structure of the brain. A recent review \citep{bernal2017deep} suggests that 3D models outperform 2D models on several neuroimaging tasks. This increase in performance, however, comes at the cost of increased computation. The computational complexity involved in applying CNNs to 3D volumes of the brain makes it difficult to explore architectures even on small datasets. In \cite{cole2017predicting}, a 10-layer 3D CNN took 83 hours to train on only 2000 images.

Existing applications of 3D convolutions to neuroimaging data \citep{cole2017predicting, cao2016mental} use architectures based on existing 2D CNN architectures. However, choices like i) the size and number of convolutional filters, ii) the type and size of pooling, and iii) the relationship between pooling, convolutional and fully connected layers, are among a myraid choices that, while extensively studied for natural images, have not been well studied in the context of 3D structural neuroimages.

Architecture design choices are often inspired by known invariances in the data. The types of invariances in natural images are typically affine, as evidenced by the data augmentation techniques, such as translations and reflections, used in existing applications of CNNs \citep{krizhevsky2012imagenet}. However, it is unclear which of these invariances arise in neuroimaging data. Given that structural brain images are aligned, translation invariance may not be as important in brain images as it is in natural images. In fact, if local patterns in the brain are location-dependent, then standard convolutions may be inappropriate. 

In this work, we begin to explore CNN architectural choices in the context of structural neuroimaging data. We propose two modifications to existing CNN architectures. Our modifications produce a network that is able to better learn patterns from brain images and trains faster than architectures developed on natural image tasks. Furthermore, these modifications are straightforward to implement in Tensorflow \citep{abadi2016tensorflow}. By sharing our implementation\footnote{\url{https://gitlab.eecs.umich.edu/mld3/brain_age_prediction}}, we hope to inspire other researchers to continue to challenge current assumptions. We summarize both the technical and clinical significance of our work below.

\paragraph{Technical Significance}
Based on the structure of neuroimaging data, we present two modifications to existing CNN architectures: 1) learning different parameters in different regions of the brain, and 2) applying more filters in the early stages of the network. These modifications enable learning distinct patterns in different brain regions. By designing these modifications to target 3D brain images, we demonstrate improved performance and training times compared to a 3D CNN baseline when predicting age from brain scans. These improvements are robust to changes to the amount of training data and the number of network parameters. Our work suggests a greater space of improvements can be achieved by redesigning CNNs for brain images.

\paragraph{Clinical Relevance}
Neuroimages are a rich form of data from which a wealth of clinically relevant labels can be predicted.  We focus on the task of predicting age from neuroimages. Age prediction from neuroimaging data allows researchers to quantify the difference in predicted age and chronological age, which can be a powerful marker of deviation from expected aging trajectories. There are a number of psychiatric and neurological conditions that are closely linked to such age-related deviations, including attention-deficit/hyperactivity disorder \citep{Shaw19649} and Alzheimer's \citep{stern2012cognitive}. Differences in predicted and chronological age have already been shown to be reliable and heritable biomarkers for a variety of neurodegenerative conditions \citep{cole2017predicting}. In addition, age prediction from brain scans enables further investigation into mechanistic factors correlated with accelerated or decelerated brain aging and associated changes in cognitive function. Our work takes the first step toward designing deep learning models that capture the invariances present in structural brain scans. We demonstrate that our ideas improve age prediction from these scans, but we hypothesize that the proposed ideas could generalize to other neuroimaging tasks.

\paragraph{} The rest of this paper is organized as follows. Section \ref{sec:RelatedWork} presents related work, both in developing CNNs on neuroimaging data and in generalizing CNNs for data with different types of invariances. Section \ref{sec:Methods} discusses the proposed architectural modifications in more detail. Section \ref{sec:ExperimentalSetup} details the dataset we use for evaluation, and our experimental setup. In Section \ref{sec:Results}, we present an empirical evaluation of our proposed approach as well as several follow-up analyses. 

\section{Related Work}\label{sec:RelatedWork}
There is a broad scope of work in applying CNNs to brain images. The most common tasks include anomaly detection, which covers tumor and micro-bleed detection, segmentation, which includes skull stripping and tumor segmentation, and label prediction. Label prediction is involves disease prediction, as well as the task we focus on: age prediction.

Predicting age from brain images is an important step towards understanding typical brain development and further understanding developmental disorders \citep{dosenbach2010prediction}. While there has been a lot of progress on the task of developing machine learning techniques for predicting age from brain images \citep{franke2012brain}, here we focus primarily on those works that utilize CNNs. \cite{cole2017predicting} propose a CNN architecture with repeated blocks of 3D, $3\times3\times3$ convolutions. Their focus is not on comparing different CNN architectures, but rather comparing the performance of their proposed architecture to a Gaussian Process model.

Few works exist that explicitly attempt to adapt CNNs to brain image tasks like age prediction or disease prediction. \cite{meszlenyi2017resting} and \cite{kawahara2017brainnetcnn} propose novel architectures to predict mild cognitive impairment and age, respectively, from functional connectivity networks extracted from functional MRI. They model the network as a graph and apply convolutions across the edges. Neither consider applying CNNs directly to brain scans. \cite{zheng2017novel} introduce a method involving a new pooling operation to predict HIV and ADHD from directly from functional imaging. However, their method requires an ensemble of CNNs and a separate algorithm to mix the ensemble, which increases computational complexity during training. Many of the works that focus on learning from brain data do not discuss architectural modifications. As described above, those that do either focus on learning from functional connectivity networks rather than actual images, or require computationally expensive models.

More generally, there is a large body of work that aims to tailor CNNs to data with different structural properties. \cite{dieleman2016exploiting} introduce four new operations to help CNNs learn invariance to rotations. \cite{gens2014deep} and \cite{cohen2016group} attempt to generalize CNNs to arbitrary groups of affine transformations. \cite{ngiam2010tiled} modify convolutions in a way that helps CNNs learn invariances present in the training data. These works focus on natural images. None consider 3D spatial data. In contrast, we investigate a different and relatively unexplored class of images. We propose a novel architecture based on the structure of 3D brain volumes. Our approach increases performance without sacrificing computational efficiency, and is straightforward to implement.  

\section{Methods}\label{sec:Methods}
This section describes two proposed modifications to existing architecture choices. The first, regional segmentation, is motivated by the idea that a CNN should learn different parameters in different regions of the brain. We segment the brain into consecutive regions and treat each region as an input channel. This encourages the network to learn region-dependent patterns. However, region-specific information is lost after the initial convolutional layers. To address this, the second modification applies more filters earlier in the network and fewer filters later. This is in contrast to the existing practice of learning fewer filters early on in the network and more in the later layers. Combined, these two modifications both improve performance on age prediction and decrease training time.

\subsection{Regional Segmentation}
CNNs are designed to capture spatially invariant patterns \citep{lecun1998gradient}. This model works well for natural images, since most objects in images retain their identity regardless of where they appear in an image. In contrast, brain images are typically aligned: each brain in a given dataset occupies the same spatial region. Complete spatial invariance is thus unnecessary since the network is not required to deduce the location of objects of interest. Furthermore, across different regions of the brain, the same pattern may have different meaning. Requiring a network to learn the same parameters over the entirety of the brain image may force those parameters to be too general and lose region-specific information. 

To better capture region-specific information, a CNN should learn different parameters in different brain regions. To address this, we propose the following modification: before applying convolutions to brain images, divide the brain image into distinct regions and concatenate those regions along the channel dimension. Learned weights will not be shared across regions, allowing the network to learn region-specific patterns.

Given that we aim to learn different patterns in different regions, how should these regions be chosen? It may seem appealing to segment the brain into known anatomical regions. However, following known anatomical regions is challenging. Anatomical regions in the brain may vary in size and shape, but it is computationally desirable to have regions with equal dimensions. Formatting these regions to share the same dimensions may lead to large quantities of zero padding and wasted computation.

\begin{figure}[htbp]
  \centering
  \begin{adjustbox}{center}
  \includegraphics[width=5in]{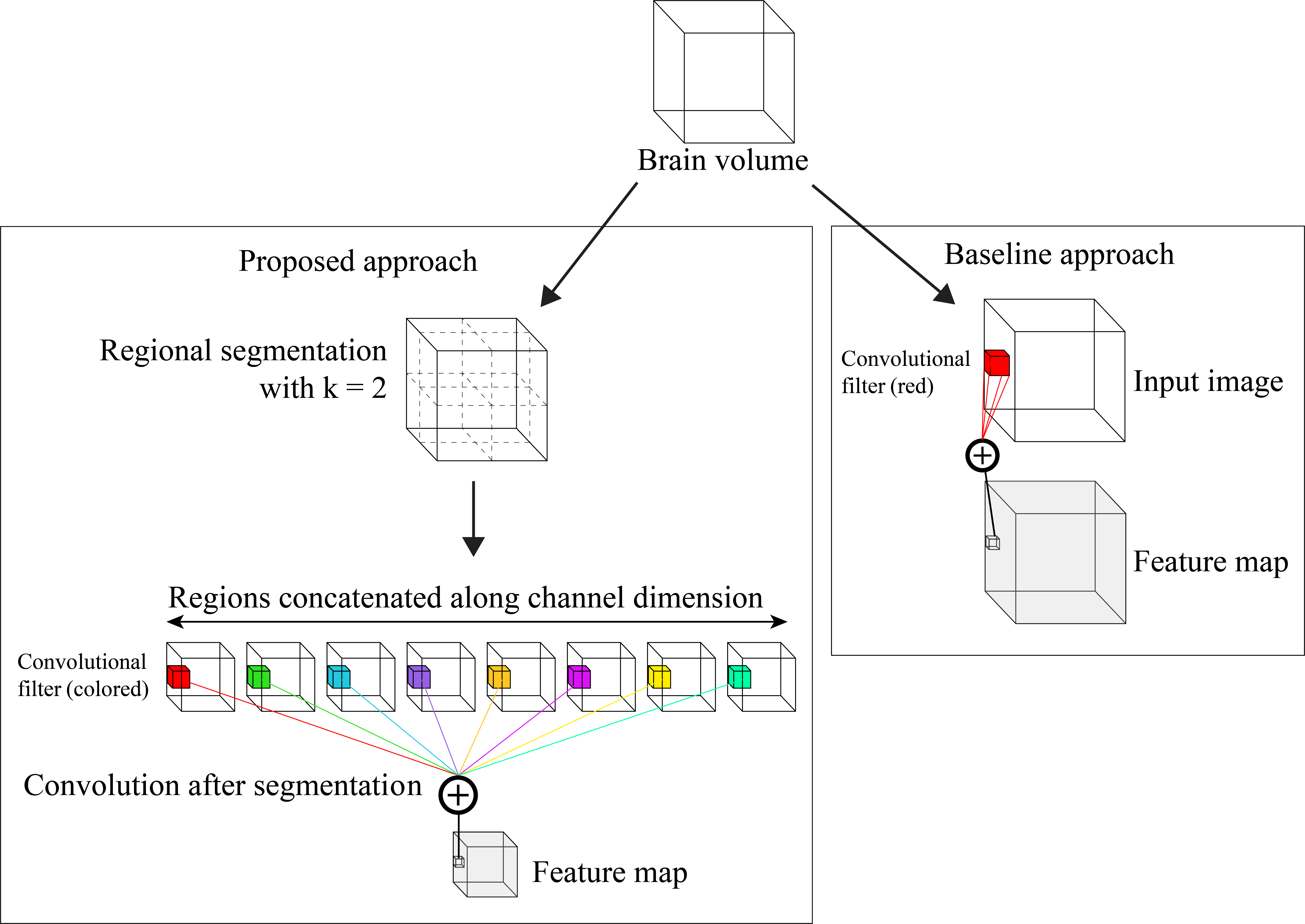} 
  \end{adjustbox}
  \caption{A depiction of regional segmentation. The operation is simple: it divides an input volume into separate, consecutive regions and then treats each region as a channel. Different colors in the figure indicate different parameters. The convolutional filter does not share parameters across regions, but applies the separate parameters simultaneously over separate regions. }
  \label{fig:Segmentation} 
\end{figure}

In light of the difficulty of using expert-defined anatomical regions, we segment the brain into consecutive, equally-sized cubes. More formally: let $I$ be a 3D, single-channel volume with dimensions $(X, Y, Z)$. Let $k$ be an integer regional segmentation rate, a hyperparameter. Regional segmentation divides $I$ into $k^3$  adjacent regions of size $(\lfloor \frac{X}{k} \rfloor, \lfloor \frac{Y}{k} \rfloor, \lfloor \frac{Z}{k} \rfloor)$. The regions are then concatenated along the channel dimension. In practice, each region is given an additional 3-voxel boundary in each dimension to avoid losing information along the edges of the regions. This approach, depicted in Figure \ref{fig:Segmentation} is easy to implement, requires no prior knowledge, and allows for different levels of granularity. Setting $k$ to be large corresponds to a finer segmentation of the brain, with less spatial parameter sharing; conversely, setting $k$ to be small corresponds to more spatial parameter sharing. Existing architectures set $k = 1$, while fully connected networks set $k$ to be the input dimension, treating each voxel as independent of the others. 

\begin{figure}[htbp]
  \centering
  \begin{adjustbox}{center}
  \includegraphics[width=4in]{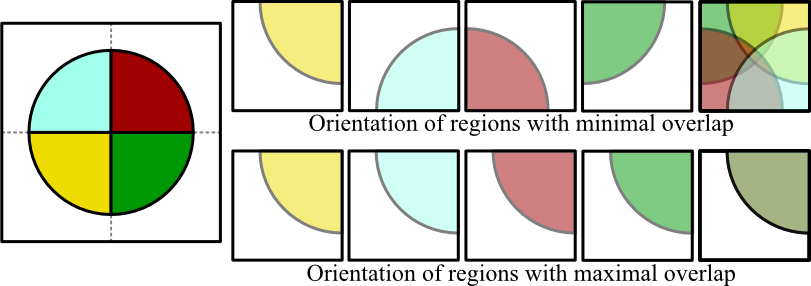}
  \end{adjustbox}
  \caption{ An illustration of the overlap in the channel dimension between regions of a center-aligned image after regional segmentation with $k=2$. The original image is depicted on the left. The top pathway depicts the default rotation of the regions, which minimizes spatial overlap. The bottom pathway depicts the rotation that maximizes spatial overlap. For illustration purposes, this image depicts a 2D circle, but the same principle applies for 3D brain scans. }
  \label{fig:RotationRegions} 
\end{figure}

Separate parameters are learned for each region since regions are no longer spatially connected. Given that these regions are now treated as separate input channels, depthwise convolutions may be more appropriate than standard convolutions, because they peform completely separate convolutions across channels \citep{chollet2016xception}. However, current deep learning libraries lack implementations of 3D depthwise convolutions, and non-native implementations are computationally expensive. Normal convolutions do learn different parameters over different channels, but combine information across channels after convolution. We mitigate this problem in two ways. First, we rotate regions into the orientation that minimizes spatial overlap across the channel dimension as depicted in Figure \ref{fig:RotationRegions}, ensuring that as much region-specific information as possible is retained during convolutions. Second, we alter the number of filters in the network to focus on the earlier layers, as discussed in the next section. 

\subsection{Filter Layouts}
Existing CNN architectures have a common trend: the number of filters in convolutional layers starts small and increases with each layer. This models human intuition: at finer detail, objects share local patterns like edges and corners. But with increasing abstraction, objects become more and more distinct. Having a large number of specialized filters in later layers of a CNN is thus critical for an object recognition task with a large number of distinct classes \citep{zeiler2014visualizing}. 

The input to our network, however, consists of segmented regions of the brain concatenated across the channel dimension. This poses a problem: the first convolutional operation across the input image combines information from all regions. With each convolution, we lose information distinguishing regions. To mitigate this issue, we propose reversing the traditional filter scheme: instead of starting with a small number of filters and increasing them, start with a large number of filters and decrease that number in later layers. We hypothesize that it is critical to extract as much information as possible from earlier stages of the image, and less important to learn information in the later layers, when information from all regions is blended together.

Our approach forces the network to learn more parameters at the lower levels, since these levels contain more information from distinct regions. This approach also bottlenecks the representation of these regions at later layers in the network, which acts as a form of regularization. 

Usually, it is difficult to experiment with a large number of filters early on in a network, because earlier filters are applied to larger images. Loading more filters on larger images results in reduced training speed and increased memory demands. However, by first applying regional segmentation and then reversing the filter layout, we find that the training time is still decreased relative to a baseline, as discussed in Section \ref{sec:Results}.

\section{Experimental Setup \& Baselines}\label{sec:ExperimentalSetup}
In this section, we describe the brain image dataset and task we used to evaluate our proposed modifications. We detail our exclusion criteria, the baseline we compare our proposed method to, and the specifics of how our models are trained. We also describe the architectural details of both the baseline and our proposed approach. 

\subsection{Dataset \& Task}
We evaluate our modifications using data from the Philadelphia Neurodevelopmental Cohort (PNC) study \citep{satterthwaite2014neuroimaging}. The PNC study is a population-based sample of children ages 8-21 ($\mu$=15.48, $\sigma$=3.19) living in the greater Philadelphia area. Neuroimaging data were collected from 1,445 subjects. Data from 997 of those subjects were made publicly available. These data were initially downloaded and preprocessed for the purposes of functional connectivity analysis. Therefore, data quality inclusion criteria follow functional connectivity guidelines (low motion, at least 4min usable data). Data from 724 of the publicly available subjects met these criteria and were included. For our analysis, we use T1-weighted structural images (voxel size $0.94 \times 0.94 \times 1$, FOV dimensions $196 \times 256 \times 160$). These images were preprocessed using the DARTEL toolbox in SPM8. Preprocessing steps included: bias field correction, brain extraction, and spatially normalizing to MNI152 template space. After preprocessing, the structural images have dimensions $121 \times 145 \times 121$. 

From these images, we aim to predict subject age. To learn a model mapping 3D images to age, we randomly split the subjects into training (n=524), validation (n=100) and test (n=100) sets. Validation data were used for model selection and stopping criteria. All reported performance metrics pertain to the test set unless otherwise specified.

\subsection{Models}\label{sec:Models}
\begin{figure}[htbp]
  \centering 
  \includegraphics[width=6in]{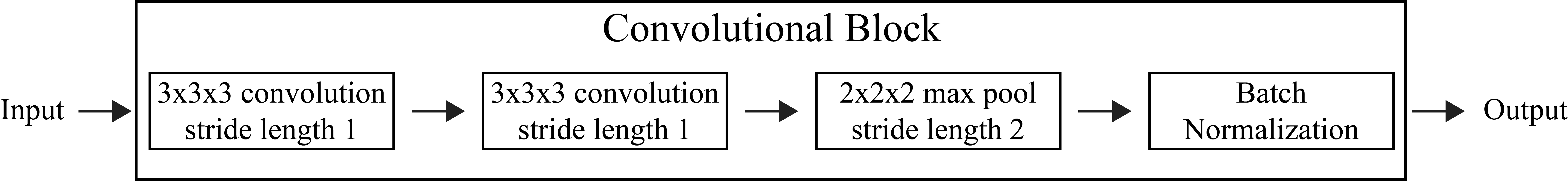} 
  \caption{ \small Both the baseline and proposed method use repeated copies of the convolutional block above. The convolutional layers within a block always have the same number of filters. }
  \label{fig:Block} 
\end{figure}

We compare a standard baseline architecture from \cite{cole2017predicting} to an architecture with our proposed modifications. Both the baseline and our proposed network are made up of repeated blocks of (convolution, convolution, batch normalization, pooling), as depicted in Figure \ref{fig:Block}. All architectures consist of 4 blocks total, followed by a hidden layer with 256 units, and then a single unit to output age prediction. All convolutional and hidden layers are followed by the eLU activation function. All models use $3 \times 3 \times 3$ convolutions of stride length 1. Before convolution, the inputs are padded so that the dimensions of the outputs are equal to the dimensions of the inputs, \textit{e.g.} ``same'' padding in Tensorflow.  Pooling layers are all $2 \times 2 \times 2$ max pooling with stride length 2. 

\begin{table}[htbp]
  \centering 
   \caption{ \small Architectures of the baseline and proposed CNNs, from left to right. Blocks 1 through 4 are copies of the block above. Both architectures have a single hidden layer with 256 units. The hidden layer uses the same activation function as the convolutional layers, namely eLu. }
      \begin{adjustbox}{center}
      \scalebox{0.8}{
  \begin{tabular}{|c|c|c|c|c|c|c|c|c|}\hline
    &&&
    \multicolumn{4}{c|}{\textbf{\# Filters}} & &  \\
    \cline{4-7}
    \textbf{Approach} &
    \textbf{Segmentation} & 
    \makecell{\textbf{Input} \\ \textbf{Channels}} &
    \textbf{Block 1} & \textbf{Block 2} & \textbf{Block 3} & \textbf{Block 4} &
    \makecell{\textbf{Hidden} \\ \textbf{Units}} &
    \makecell{\textbf{Output} \\ \textbf{Units}} \\
    
     \hline
    Baseline & No & 1 & 8 & 16 & 32 & 64 & 256 & 1\\ \hline
    Proposed & Yes & $k^3$ & 64 & 32 & 16 & 8 & 256 & 1 \\ \hline
  \end{tabular}
  }
  \end{adjustbox}
  
  \label{tab:Arch} 
\end{table}

The \textit{Baseline} architecture has 8 filters in its first block of convolutions, and double that number in every successive block, for a total of 64 filters in the final block. Our \textit{Proposed} approach has 64 filters in its first block, and half that every block after, for a total of 8 filters in the final block.  A summary of both architectures is presented in Table \ref{tab:Arch}. Each layer of convolutional filters can be thought of as a 5-D tensor of shape $C \times 3 \times 3 \times 3 \times M$, where $C$ is the output dimension of the previous layer, and $M$ is the number of filters (the output dimension of the current layer). For example, in the first layer of the baseline network, the original input image is convolved with a 5-D tensor of shape $1 \times 3 \times 3 \times 3 \times 8$, while in the first layer of the proposed network, the segmented input image is convolved with a 5-D tensor of shape $k^3 \times 3 \times 3 \times 3 \times 64$.

\subsection{Implementation Details}\label{sec:implementation}
For all experiments, every model is trained five times with different random initializations until performance on the validation set no longer improves, approximately 700 epochs. Weights are initialized using Xavier initialization \citep{glorot2010understanding}. For each run, the model weights that achieved the best performance on the validation set during training were saved. These weights were restored after each run and the model was evaluated on the test set. We report the average test performance (and standard deviation) over the different initializations. 

For computational efficiency, all images are pooled by a factor of 3 in each dimension before being fed into the network, resulting in images of size $41 \times 49 \times 41$. We pool the images by averaging consecutive $3 \times 3 \times 3$ blocks; the specific aggregation function performed is discussed in the next section.

We train all models with the Adam optimizer \citep{kingma2014adam}. We found no significant performance improvement for any model tested with non-default parameters, so we keep the defaults ($\alpha=0.001, \beta_1=0.9, \beta_2=0.999, \epsilon=10^{-8}$). We optimize with respect to mean squared error (MSE) on the training set. We report both MSE and mean absolute error (MAE) on the held-out test set. We implemented all models using Tensorflow \citep{kingma2014adam} and ran all experiments on a GeForce GTX 1080 Ti GPU. 

To select the hyperparameter $k$ for our proposed model, we searched in the range $[1,2,3,4]$, where 1 corresponds to no segmentation to and 4 corresponds 64 regions. $k$ was chosen based on the model that provided the lowest average performance (MSE) on the validation set. 

\section{Results \& Discussion}\label{sec:Results}
In this section, we compare the performance of our proposed approach  to the baseline architecture described in Section \ref{sec:Models}. We also compare the performance of each proposed modification independently. We measure performance in terms of both error and training time. We present an in-depth follow-up analysis, in which we investigate performance across a range of different settings.

%\textcolor{red}{For the record, the performance of the proposed-without-reverse model is below. One reviewer thought we should include this information to prove the validity of the \textit{reverse filters} idea.}
%\begin{table}[htbp]
  %  \centering
  %  \begin{tabular}{|c|c|c|} \hline
   %     \textbf{k} & \textbf{MSE} &  \textbf{MAE} \\ \hline
   %      2 & $3.76 \pm 0.08$ & $1.59 \pm 0.03$ \\ \hline
   %      3 & $3.92 \pm 0.34$ & $1.61 \pm 0.03$ \\ \hline
    %     4 & $3.84 \pm 0.66$ & $1.59 \pm 0.16$ \\ \hline
   % \end{tabular}
  %  \caption{Performance of standard filter direction model.}
  %  \label{tab:no_reverse_table}
%\end{table}

\subsection{Prediction Error}
In Table \ref{tab:PNC1} we report the MSE and the MAE for the task of predicting age from structural brain images. As described in Section \ref{sec:implementation}, reported performances below represent the average and standard deviation of five different initializations of each model evaluated on the held-out test set.

\begin{table}[htbp]
  \centering 
  \caption{\small Comparison of model performances for the task of predicting age on the PNC dataset. The regional segmentation rate $k$, as described in Section 3, refers to how finely the brain is segmented before filters are learned in each region. We find that 2 is the optimal setting for this task. The filter layout refers to either increasing the number of filters (8-16-32-64), or decreasing the number of filters (64-32-16-8), as the depth of the network continues.}
      \begin{adjustbox}{center}
  \begin{tabular}{|c|c|c|c|c|c|}\hline
    \textbf{Approach} & \textbf{Filter Layout} & \textbf{$k$} & \textbf{MSE} & \textbf{MAE} & \textbf{\makecell{Training Time\\ (minutes)}} \\ \hline
    Baseline & 8-16-32-64 & - & $3.72 \pm 0.20$ & $1.58 \pm 0.06$ & 52.5\\ \hline
    Proposed & 64-32-16-8 & 2 & $\mathbf{3.03 \pm 0.15}$ & $\mathbf{1.43 \pm 0.03}$ & 40 \\ \hline
    Segmentation only & 8-16-32-64  & 2 &  $3.76 \pm 0.08$ & $1.59 \pm 0.03$ & \textbf{30} \\ \hline
    Reverse layout only & 64-32-16-8  & - &  $3.64 \pm 0.35$ & $1.53 \pm 0.05$ & 117 \\ \hline
  \end{tabular}
  \end{adjustbox}
 
  \label{tab:PNC1} 
\end{table}

\begin{figure}[htbp]
  \centering 
  \includegraphics[width=4in]{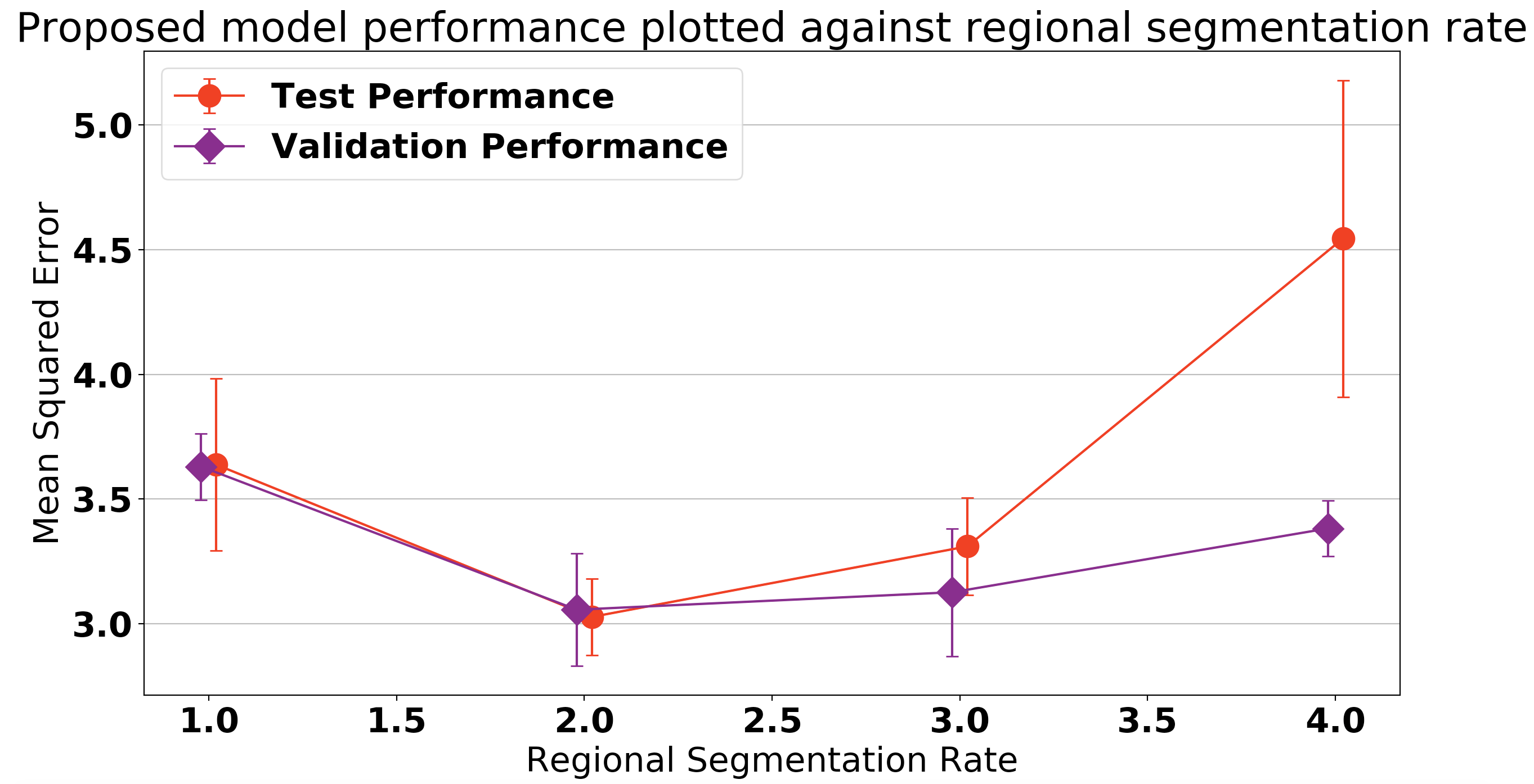} 
  \caption{\small Performance of proposed model plotted against the regional segmentation rate, $k$. The points are offset from integer values to better display both test and validation performance. Each point is the mean of five runs, and the error bars are the standard deviation across those runs. This chart demonstrates that for this particular problem, $k=2$ is a locally optimal hyperparameter. }
  \label{fig:TestKPlot} 
\end{figure} 

For the proposed model, we selected a regional segmentation rate of $k=2$ based on validation performance, resulting in eight different regions. Our experiments show that the combination of segmenting the brain into eight separate regions and reversing the order of filters results in improved age prediction from structural brain data. Results indicate that it is the combination of both modifications that improves performance; neither modification alone lowers the prediction error from the baseline.

To understand how model performance varies with $k$, we plot both the validation and test performance of our proposed method for each setting of $k$ in Figure \ref{fig:TestKPlot}. Each point is the average performance of the model across five initializations, and error bars represent the standard deviation. For our proposed architecture, segmenting the brain too finely or leaving the brain whole results in decreased performance. We only used validation performances to do hyperparameter selection, but the same trend holds for both test and validation data.

%\begin{figure}[htbp]
 % \centering 
 % \includegraphics[width=5in]{TimeBarPlot.png} 
  %\caption{Training time per 700 epochs plotted against image pooling rate. 
  %Although reversing the filters is expensive, the time benefit gained by regional segmentation is great enough that proposed model with segmentation is faster than the baseline without segmentation.}
 % \label{fig:TimeBarPlot} 
%\end{figure} 

\subsection{Training Time}
%\textcolor{red}{
Table \ref{tab:PNC1} also displays the training time of the evaluated models. Training time refers to wall clock time, in minutes, per 700 epochs, the number of epochs taken to train each model. Each model was trained on the same GPU under the same load, for fair comparison. The table shows that reversing the layout of the filters in a CNN is a computationally expensive modification. Doing so more than doubles the training time from the equivalent baseline. More filters earlier in the network results in more filters being applied when the image data is larger and convolutions are more expensive.

Table \ref{tab:PNC1} also shows that applying regional segmentation nearly halves training time. This is due in part to the reduced number of fully connected parameters in the proposed network, as discussed in section \ref{sec:fc}, and in part to the size of the feature maps in the network after the first convolution. A similar volume of data is input to the first convolution of the baseline and the segmentation only network. In the former, the images are arranged as ($41 \times 49 \times 41 \times 1$) arrays. In the latter, the images are arranged as ($23 \times 27 \times 23 \times 8$) arrays\footnote{Note that the dimensions of the segmented image are not exactly equal to ($\lfloor \frac{41}{2} \rfloor \times \lfloor \frac{49}{2} \rfloor \times \lfloor \frac{41}{2} \rfloor \times 8$) due to the padding boundary described in section \ref{sec:Methods}.}. After the first convolution, the number of feature maps in each section of the network is the same, but the maps are roughly twice as small in each spatial dimension in the segmentation only network, meaning the segmentation only network performs less computation in the remaining convolutional layers. 

The underlying implementation of the convolution operation also affects training time. Consider an input image of shape ($N \times X \times Y \times Z \times C$) where $N$ is the number of images in a batch, $X, Y, \textrm{ and } Z$ are spatial dimensions, $C$ is the number of channels in the image and a convolutional filter tensor of shape ($C \times \ell \times \ell \times \ell \times M$) where $\ell$ is the size of the convolutional filter and $M$ is the number of filters. The convolution can be thought of as a matrix multiplication between a filter matrix of shape ($M \times C\ell^3 $) and an input matrix of shape\footnote{This assumes that the output dimension is made to be equal to the input dimension using \textit{same} padding.} ($C\ell^3 \times NXYZ$). This multiplication is more efficient if the products $C \ell^3$ and $NXYZ$ are similar in magnitude, rather than one being significantly larger than the other \citep{chetlur2014cudnn}. If the input image is a single-channel, 3D brain image, and the filter size is small, the product $C\ell^3$ will be much smaller than $NXYZ$. In our baseline network, $C\ell^3 = 27$, while $NXYZ=4*41*49*41=329476$. Regional segmentation increases $C$ while decreasing $XYZ$ by a factor of $k^3$, making the products $C\ell^3$ and $NXYZ$ closer in magnitude and resulting in a more efficient convolution.

To demonstrate how regional segmentation impacts the speed of convolutions, we simulated the forward pass of the convolution operation on a randomly-generated $4\times \frac{72}{k}  \times \frac{72}{k}  \times  \frac{72}{k}  \times k^3$ image. We convolved the image with a randomly generated $k^3 \times 3\times3\times3 \times 8$ filter, similar to the first convolutional layer of our baseline architecture. We varied the segmentation rate $k$ in the range $[1,2,3,4,6,8,9,12,18,24,36]$, choosing the divisors of 72 and forgoing the voxel boundary at the edges of regions to ensure that each convolution was applied over exactly the same amount of data. Each data-point of Figure \ref{fig:SimulatedConvolutions} displays the result of timing 500 forward-pass convolutions at the indicated segmentation rate. Figure \ref{fig:SimulatedConvolutions} demonstrates that initially as the segmentation rate increases and the products $C\ell^3$ and $NXYZ$ become closer in magnitude, the convolution speeds up. But, as the image becomes too finely segmented, $C\ell^3$ exceeds $NXYZ$ and the convolution slows down.

\subsection{Robustness to Pooling Strategy}

\begin{figure*}%[H]
\centering
\begin{subfigure}{0.54\textwidth}
  \centering
  \includegraphics[width=1\linewidth]{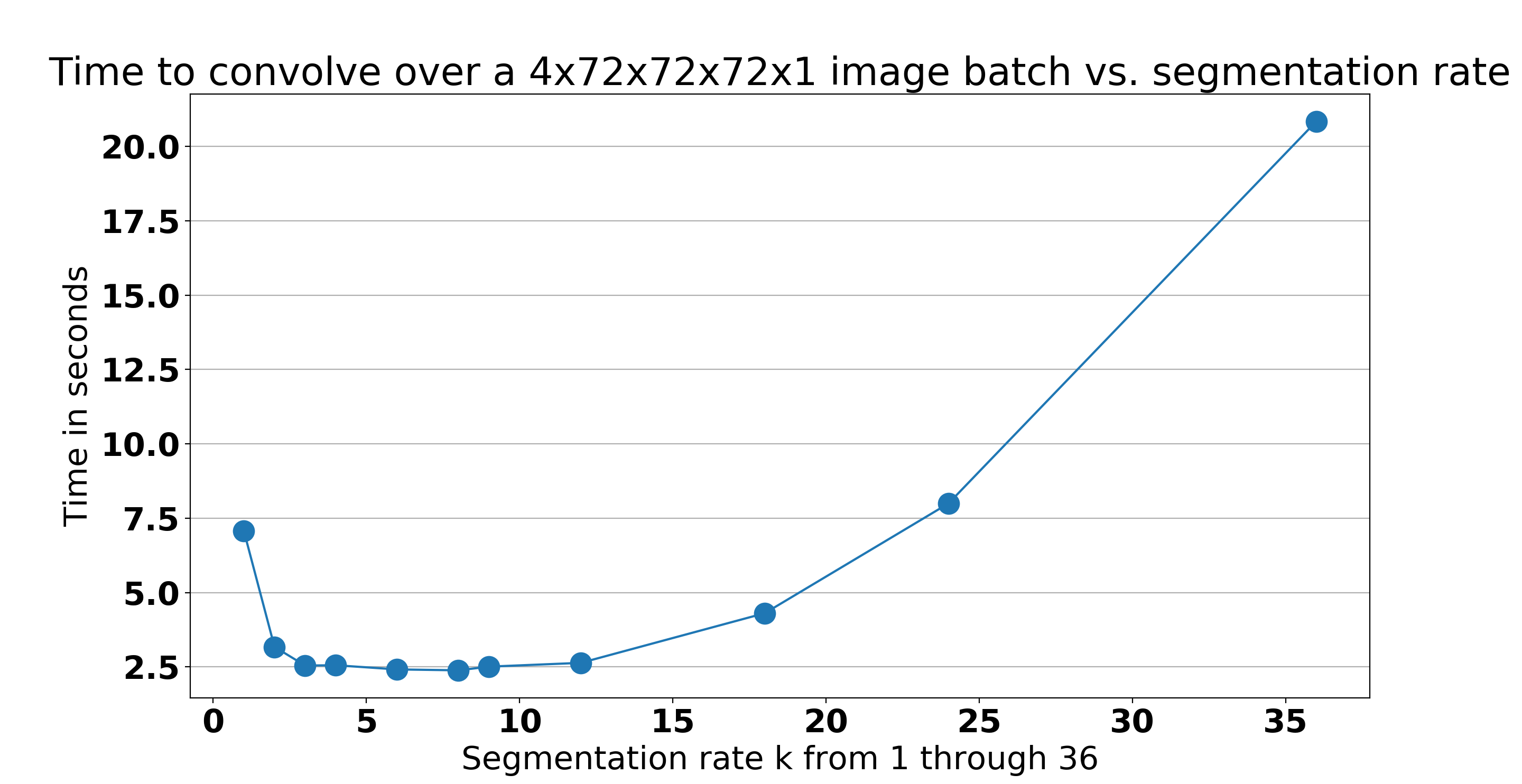}
  \caption{}\label{fig:SimulatedConvolutions}
\end{subfigure}
\begin{subfigure}{0.45\textwidth}
  \centering
  \includegraphics[width=1\linewidth]{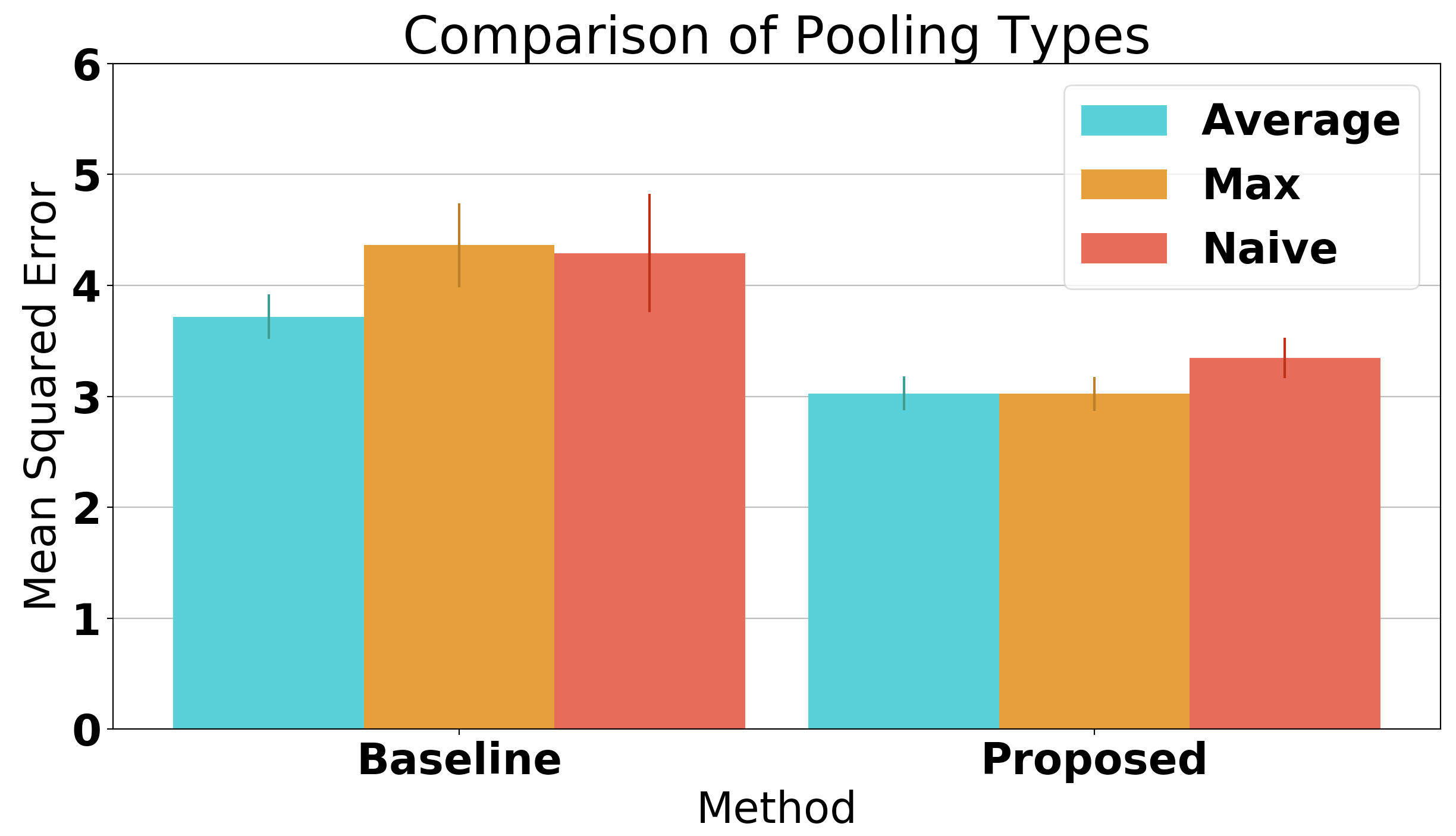}
  \caption{} \label{fig:PNC_pooling_strategy} 
\end{subfigure}

\caption{\small Convolution time vs. segmentation rate on the left, and performance based on pooling type on the right. \textbf{(a)} Time to compute the forward pass of a convolution on a 3D, single-channel image batch after regional segmentation. The plot shows that segmenting the image decreases training time up to a point, but too finely segmenting the image results in increased training time. \textbf{(b)} Plot of performance on pooling types. \textit{Average} and \textit{Max} refer to average and max pooling, respectively, while \textit{Naive} refers to downsampling the image by keeping the first voxel in every consecutive $3 \times 3 \times 3$ block.  Average pooling is the best pooling type for both proposed and baseline. The proposed model outperforms the baseline using any pooling type, and is less sensitive to how the images are pooled. }
\label{fig:TimeAndPooling}
\end{figure*}

In both the baseline and proposed approach, we average pool the images before feeding them into the network to speed up training. Average pooling smooths out signals within the original images, but other pooling strategies may have different properties. To better understand how this smoothing affects the relative performance of to our proposed approach to the baseline, we evaluate both models on two other pooling strategies. \textit{Max} pooling corresponds to taking the maximum value of every consecutive $3 \times 3 \times 3$ block. \textit{Naive} pooling refers to taking the first voxel and discarding the rest in every consecutive $3 \times 3 \times 3$ block. While average pooling smooths signals, max pooling emphasizes strong signals and naive pooling arbitrarily selects voxels based on order. Given these differences, we hypothesized that relative performance may vary based on the pooling type. To test this hypothesis, we train the baseline and our proposed model separately on the training images pooled with each strategy. We evaluated the MSE of each model on the held-out test set pooled with each strategy, respectively (Figure \ref{fig:PNC_pooling_strategy}).

Average pooling is the best strategy for both the baseline model and our proposed method. Our proposed method, however, is more robust to pooling strategies than the baseline. Most importantly, our proposed method reports a lower MSE for all pooling strategies than the baseline for any pooling strategy. 

%\begin{figure}[htbp]
 % \centering 
 % \includegraphics[width=5in]{TwoSampleComp.png}   \caption{Plot of performance on pooling types. \textit{Average} and \textit{Max} refer to average and max pooling, respectively, while \textit{Naive} refers to downsampling the image by keeping the first voxel in every consecutive $3 \times 3 \times 3$ block.  Average pooling is the best pooling type for both proposed and baseline. The proposed model is less sensitive to how the images are pooled. }
 % \label{fig:PNC_pooling_strategy} 
%\end{figure}

\subsection{Sensitivity to Number of Fully Connected Parameters}\label{sec:fc}
Our proposed model and the baseline model have a similar number of convolutional parameters (233,280 compared to 219,672). However, because the proposed model has fewer convolutional filters in later layers, it also has fewer parameters in the fully connected layers, despite having the same number of hidden units. Specifically, the proposed model has 16,640 weights in its fully connected layers, while the baseline has 590,080 such weights, despite both having 256 hidden units. In order to examine how much the number of fully parameters impacts performance, we varied the number of units in the hidden layer in our proposed approach in the range $[256, 2496, 4736, 6976, 9216]$. We varied the number of hidden units in our baseline in the range $[7, 69, 131, 194, 256]$. These ranges were chosen because they result in a similar number of fully connected parameters for each model, within 1\% of difference. Note that $256$ is the default number of hidden units for both models. 

The resulting performance of each model across these different settings is plotted in Figure \ref{fig:PNC_fc_param}. Having fewer fully connected parameters hurts the baseline. In contrast, the reverse model remains within the margin of error regardless of the number of fully connected parameters. These results suggest that the difference in performance between the proposed approach and the baseline is not due to the proposed approach having fewer fully connected parameters. 

%\begin{figure}[htbp]
 % \centering 
 % \includegraphics[width=4in]{FCPlot.png} 
 % \caption{Plot of performance vs. number parameters in the fully connected layers. On the left side, at 16,640 parameters, is the default for the proposed model. On the right side, at 590,080 parameters, is the default for the baseline model. We vary the number of fully connected parameters by changing the number of units in the fully connected layer. Reducing the number of parameters for the baseline model does not help its performance: rather, it hurts it. }
 % \label{fig:PNC_fc_param} 
%\end{figure}

\subsection{Varying the Amount of Training Data}
Lack of data is a large concern in MRI studies, because gathering research-grade MRI scans is expensive and time consuming. It is not uncommon for researchers to work with datasets of fewer than 200 patients \citep{zheng2017novel, meszlenyi2017resting}. Compared to other single-site datasets, the PNC dataset is relatively large. In our setup, we have 524 training examples. To better understand the impact of small training sizes on our conclusions, we varied the number of training examples in the range $[100, 200, 300, 400, 524]$. For each number of training examples, we independently and randomly selected examples without replacement from the original training set of 524.

\begin{figure*}%[H]
\centering
\begin{subfigure}{0.48\textwidth}
  \centering
  \includegraphics[width=1\linewidth]{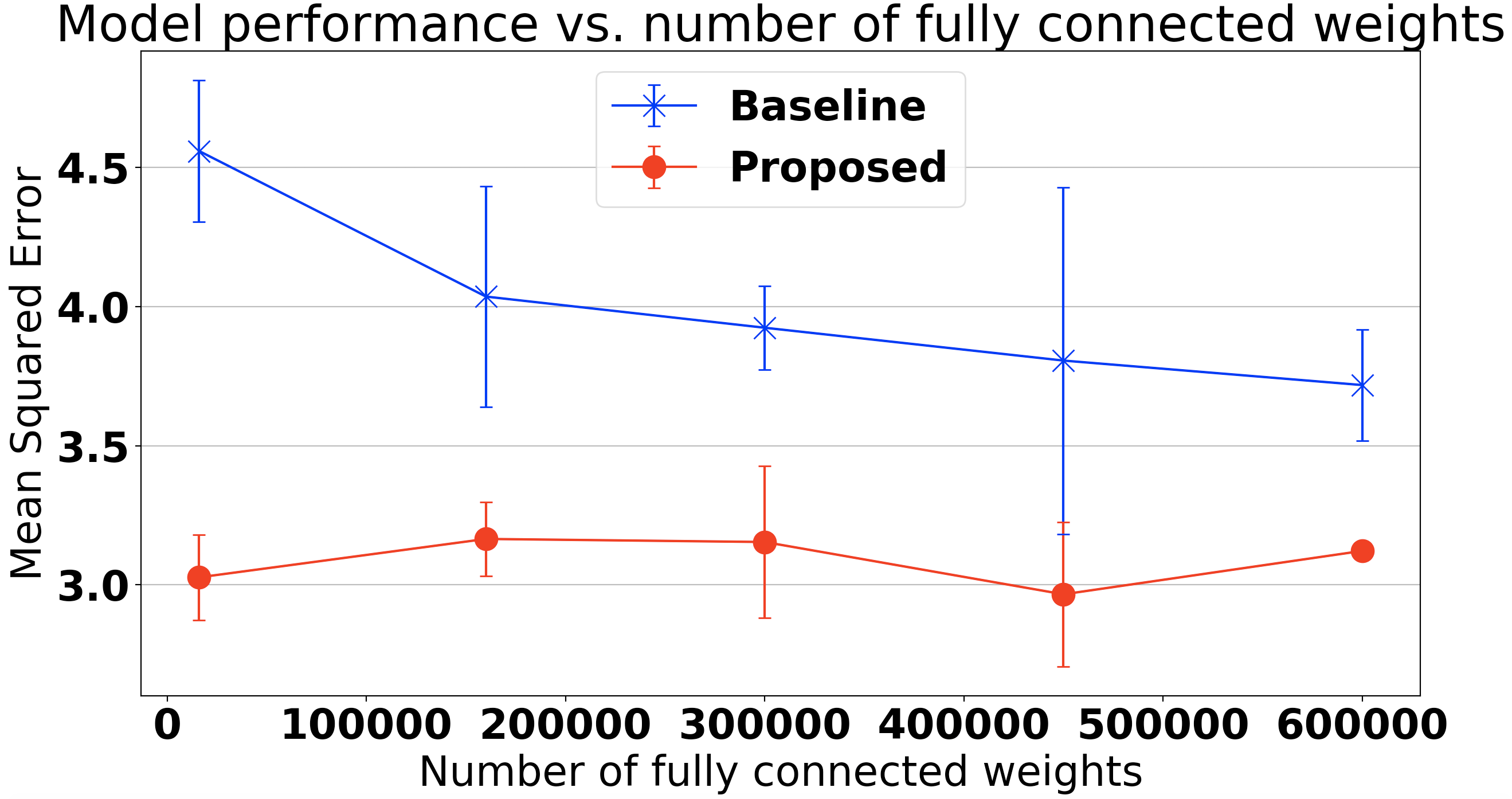}
  \caption{} \label{fig:PNC_fc_param} 
\end{subfigure}
\begin{subfigure}{0.48\textwidth}
  \centering
  \includegraphics[width=1\linewidth]{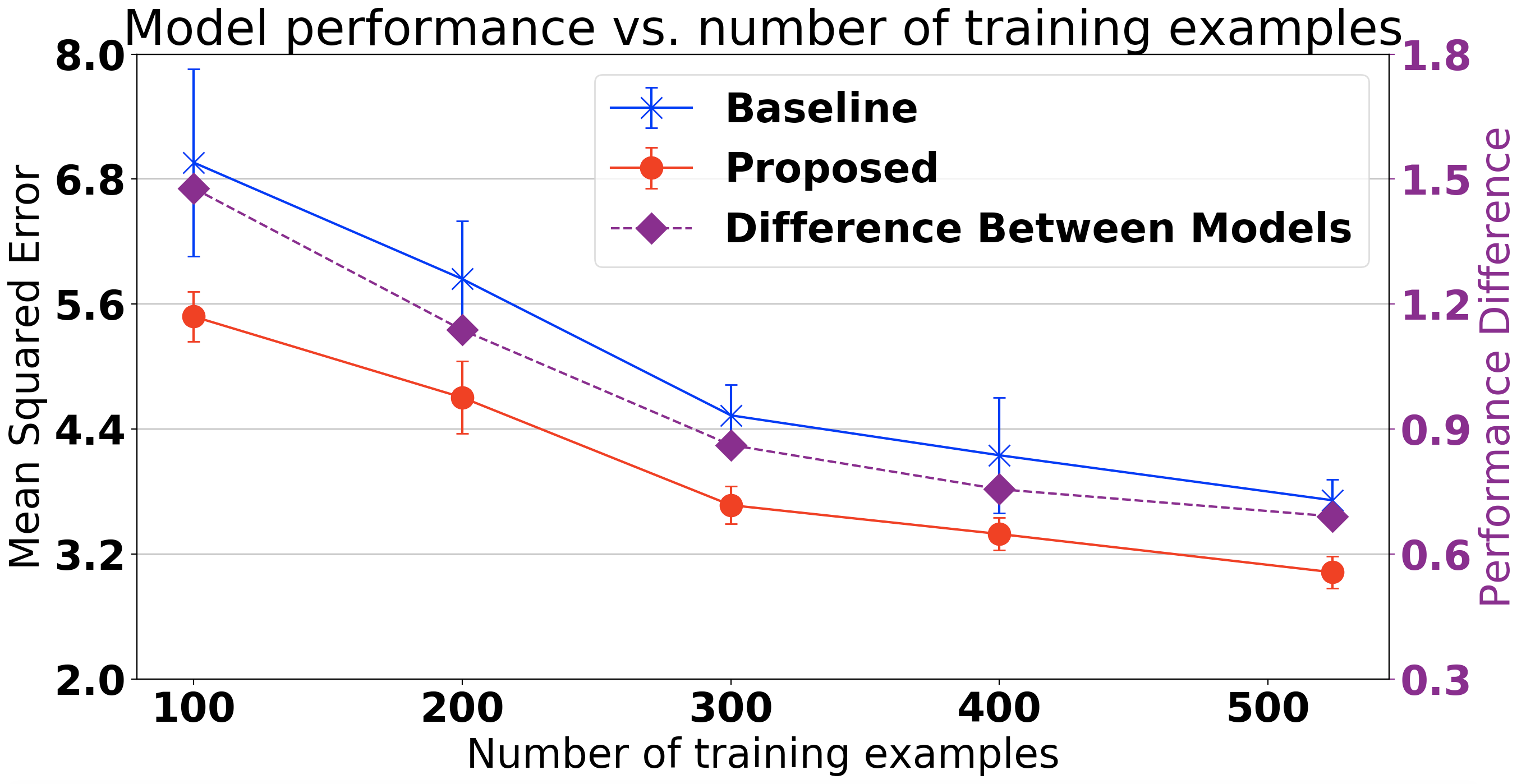}
  \caption{} \label{fig:PNC_training_set_size} 
\end{subfigure}

\caption{\small Investigating the robustness of observed improvements \textbf{(a)} Plot of performance vs. number parameters in the fully connected layers. The left side, 16,640 parameters, is the default for the proposed model. The right side, 590,080 parameters, is the default for the baseline model. We vary the number of fully connected parameters by changing the number of units in the fully connected layer. Reducing the number of parameters for the baseline model does not help performance: rather, it hurts it.  \textbf{(b)} Plot of performance vs. number of examples in training set size. The full training set, 524 examples, is on the right. The performance gap between our model and the baseline increases as training examples decrease. }
\label{fig:ParamsAndTraining}
\end{figure*}

In Figure \ref{fig:PNC_training_set_size}, we plot the number of training examples vs. model performance in terms of MSE for both the baseline and proposed approach. The line in purple plots the absolute difference in performance between the two models. The performance of both approaches declines as we decrease the number of training examples, as expected. However, the difference between the two models increases as the number of training examples decreases. This suggests that our method is more robust to small training sets. Furthermore, our proposed approach exhibits less variation in performance, indicating that it is potentially less more robust to weight initialization.

%\begin{figure}[htbp]
%  \centering 
 % \includegraphics[width=4in]{NumberExamples.png} 
 % \caption{Plot of performance vs. number of examples in training set size. The full training set, 524 examples, is on the right. Although we see very similar performance declines for both models when training sizes get small, the proposed approach maintains smaller error bars at low training sizes, which is important for accurate estimation.  }
 % \label{fig:PNC_training_set_size} 
%\end{figure}

\section{Conclusions}\label{sec:Conclusions}

In this paper, we introduced two novel modifications to existing CNN architectures, inspired by the structure of brain image data. First, we apply different learned parameters to different regions in the brain. Second, we start with a large number of filters in the first layer and decrease the number of filters after each convolutional block. These modifications encourage the network to learn region-specific patterns in the brain. Combined, they are simple and easy to implement, and result in a model that trains faster than a comparable CNN baseline. In addition, using our proposed architecture, we consistently observed improved age prediction across multiple downsampling types and varying amounts of training data. 

Our work suggests that there is a larger space of possible architectural choices to explore when adapting CNNs to brain imaging tasks. Due to fundamental differences between brain images and natural images, previous findings about how to build CNNs may not apply to neuroimaging data, or other types of data outside of the domain of natural images. This work provides a starting point to challenge assumptions existing architectures make when applying CNNs to brain images. 

There exists a variety of ways to expand this work. We did not focus increasing the depth of the neural network, or techniques associated with deeper networks, like inception modules \citep{szegedy2015going} or skip connections \citep{he2016deep}. However, if publicly available brain image datasets increase by an order of magnitude, such techniques may be of interest. Our work also did not focus on comparing pooling types within the network, activation functions, or larger convolution sizes. However, a better understanding of how to best design architectures for 3D brain volumes demands investigation in these areas as well.  We hope that our work inspires more questions into how to best design CNNs for neuroimaging learning tasks. Understanding how to challenge and adapt ideas from existing CNN architectures will be critical to better predicting a multitude of clinically relevant labels, from age to Alzheimer's.

% ACKNOWLEDGEMENTS ONLY GO IN THE CAMERA-READY, NOT THE SUBMISSION
\acks{This work was supported by the Exercise and Sports Science Initiative (grant no. U056408), and  the National Science Foundation (NSF award no. IIS-1553146). The views and conclusions in this document are those of the authors and should not be interpreted as necessarily representing the official policies, either expressed or implied, of the Exercise and Sports Science Initiative or the NSF.}

\bibliography{main}

\end{document}